\documentclass[conference]{IEEEtran}
\usepackage{graphicx}
\usepackage{array}

\usepackage{hyperref}

\begin{document}

\title{Using Autoencoders To Learn Interesting Features For Detecting Surveillance Aircraft}

\author{
\IEEEauthorblockN{Teresa Nicole Brooks}
\IEEEauthorblockA{
Seidenberg School of CSIS\\
Pace University\\
New York, NY\\
Email: tb93141n@pace.edu}
}
\maketitle

\IEEEpeerreviewmaketitle

\begin{abstract}
This paper explores using a Long short-term memory (LSTM) based sequence autoencoder to learn interesting features for detecting surveillance aircraft using ADS-B flight data. An aircraft periodically broadcasts ADS-B (Automatic Dependent Surveillance - Broadcast) data to ground receivers. The ability of LSTM networks to model varying length time series data and remember dependencies that span across events makes it an ideal candidate for implementing a sequence autoencoder for ADS-B data because of its possible variable length time series, irregular sampling and dependencies that span across events.
\end{abstract}

\section{Introduction}
The motivation for this research was inspired by the original research presented by Richards, MacDonald-Evoy, and Hernandez in their ``Tracking Spies In The Skies'' talk at DEF CON 25 \cite{defcon25spies}. The goal of their research is to leverage ADS-B (Automatic Dependent Surveillance - Broadcast) data that is broadcast by commercial and private aircraft to detect surveillance aircraft. Their approach for classifying surveillance aircraft uses a manually built model that was developed from their visual observations of flight patterns of known surveillance aircraft. The original interest in this research and ADS-B data was motivated by the desire to build a statistical model that could be used to implement a more robust surveillance aircraft detection system. This paper marks the first steps of implementing such a classification system. We show that it is possible to extract interesting features for surveillance aircraft by training a model using a LSTM based autoencoder and examining the network's encoder layer output.

The remainder of this paper is organized as follows. Section II provides background information.  In Section III, is a brief summary of related work. Section IV, documents the experiment design. In Section V, we discuss the minimum feature set extracted for positive training samples. The last sections contain proposed future work, network architecture and conclusions.
\section{Background}

This section provides a brief discussion of relevant subject matter topics.

\subsection{Automatic Dependent Surveillance - Broadcast (ADS-B)}
ADS-B data is surveillance data that an aircraft periodically broadcasts to ground receivers. This data allows an aircraft to be tracked and is often used by air traffic control stations as well as other aircraft. ADS-B data is not encrypted and can be read by any receiver that is tuned to the 1090 MHz frequency. The public availability of this data makes it attractive for use by enthusiasts and researchers. There are two options for getting access to this data, one is to use a software defined radio (SDR) to implement your own receiver or use real-time data collected from crowdsourced ADS-B data websites such as ADS-B Exchange \cite{adsbexchange} and Flightrader24 \cite{radar24}.

\subsection{Autoencoders}
Autoendcoders are special neural networks that can be used as a pre-processing step in machine learning systems to reduce dimensionality of a given dataset. This property of autoencoders makes them ideal for unsupervised learning of features vs manually choosing features based on intuition or trial and error. Essentially they work by copying the input data to the output, but this copy is not an exact replica. The copy is a compressed version of the data, this restriction forces autoencoders to keep only the most interesting features.

\section{Related Works}

This section provides a brief summary of related works for this project.

\subsection{The Use Of Autoencoders For Discovering Patient Phenotypes}
In The Use Of Autoencoders For Discovering Patient Phenotypes \cite{Suresh2017} researchers explore the use of autoencoders to discover patient phenotypes. It is their hypothesis that the interesting features extracted during this process could be used to make predictions on the onset and weaning patients off of treatments (interventions) \cite{Suresh2017}. The over all goal of this research is to ultimately provide higher quality care by enabling physicians to make more informed, data driven decisions by leveraging machine learning techniques and the availability of electronic health records. 

In this paper they compared the performance of autoencoders that took fixed length sequences of concatenated time steps as input vs recurrent (LSTM) sequence to sequence autoencoders. They also evaluated the performance of various configurations of these autoendoders by measuring the mean squared error (MSE) between the predicted sequence of values and the true sequence of values.

Overall the results are promising and show that a single layer LSTM sequential autoencoder can achieve a lower MSE than a single layer fixed length autoencoder. This performance holds when that data is constructed as various length timeseries as well as data stratified across multiple care units at a fixed timeseries intervals of 32 hours.

\subsection{Flight Phase Identification from ADS-B Data Using Machine Learning Methods}
In Flight Phase Identification from ADS-B Data Using Machine Learning Methods \cite{Sun2016} researchers explores identifying phases of flight from ADS-B data by applying a unsupervised learning technique called clustering to bucket flights into full or partial flight paths. They then perform further segmentation on this data, by applying fuzzy logic to the full and partial flight paths found via the initial clustering step. The overall goal of this research is to use their findings to build open aircraft performance models and to integrate these models into open-source ATM (air traffic management) simulator called BlueSky.

The results showed that researchers were indeed able to construct labeled flight phase data for each entry in their validation data set. 

\section{Experiment Design}

\subsection{Features}
The primary purpose of this phase of research is to build a model that will extract interesting features for surveillance aircraft with the anticipation that these features can be used to train classifiers that can identify these types of aircraft by observing flight data. All the data used in this experiment was pulled from the daily historic data API provided by ADS-B Exchange. Each daily query, gives approximately 7.9 GBs of ADS-B data for all aircraft that have ``short trails'' turned on. Short trails, shows every position of an aircraft over the time span of a few seconds.

\begin{table}[h!]
\begin{tabular}{ |m{1.5cm}|m{2cm}|m{4cm}| } 
 \hline
 Field & Data Type & Description \\
 \hline
 Tsec & integer & The number of seconds that the aircraft has been tracked for.\\
 \hline
 Cmsgs & integer & The count of messages received for the aircraft. \\
  \hline
 Alt & integer & The altitude in feet at standard pressure.\\
  \hline
  Galt & integer & The altitude adjusted for local air pressure.\\
  \hline
  InHG & float & The air pressure in inches of mercury that was used to calculate the AMSL altitude from the standard pressure altitude.\\
  \hline
  Lat & float & The aircraft’s latitude over the ground.\\
  \hline
  Long & float & The aircraft’s longitude over the ground.\\
  \hline
      PosTime & epoch (ms) & The time that the position was last reported by the aircraft.\\
      \hline
  Spd & knots (float) & The ground speed in knots.\\
  \hline
  SpdTyp & integer & The type of speed that Spd represents.\\
  \hline
  Trak & degrees (float) & Aircraft’s track angle across the ground clockwise from 0° north.\\
  \hline
  TrkH & boolean & true, if Trak is the aircraft's heading\\
  \hline
  Vsi & integer & Vertical speed in feet per minute.\\
  \hline
  Gnd & boolean & true, if aircraft is on the ground\\
  \hline
  Trt & integer & Transponder type.\\
  \hline
  Talt & number & The target altitude, in feet, set on the autopilot / FMS etc.\\
  \hline
  Ttrk & number & The track or heading currently set on the aircraft’s autopilot or FMS. \\
  \hline
\end{tabular}
\\
\caption{ADS-B broadcasted data fields as per ADS-B Exchange API documentation \cite{adsbexchange}}
\label{table:1}
\end{table}

Some data fields provided by ADS-B Exchange are derived by the service itself or are calculated values not broadcasted by from aircraft's transponder. Because the goal is to later use the features learned during this experiment to classify various types of surveillance aircraft by observing flight data, we will only use natively broadcasted data points in autoencoder input feature vectors. \ref{table:1} shows a list of the 17 ADS-B fields used to build feature vectors, their corresponding data type and description.

\subsubsection{Query Filters}
For this experiment, we queried the ADS-B historic data API for ${9}$ days flight data for ${41}$ unique surveillance aircraft. We then filtered this flight data using a set of unique registration numbers of known surveillance aircraft and by filtering on an ADS-B Exchange API boolean flag which marks aircraft as ``interesting.'' Typically aircraft marked as interesting are used by law enforcement or for medical transport. For all experiments, only aircraft registered in the United States was used. 

\subsection{Pre-processing}
Before any training or inference can take place the raw ADS-B data must go through a number of pre-processing steps in order for it to be useful in any machine learning system.  This section will discuss the pre-processing steps necessary to prepare ADS-B data to be used in a LSTM based autoecoder.

\subsubsection{Random Sampling}
Typically an aircraft can broadcast its position once every second. This high broadcast rate leads to redundant data which makes it difficult to work with because of the potentially large volume of data produced per aircraft. The input to the LSTM based autoencoder will be an input vector that represents a sequence of events. Each event is represented by a feature vector of ${x}$ values. The corresponding data fields for these values are listed in \ref{table:1}.  To eliminate redundant data points, we will perform random sampling for each aircraft in the data set, this will be done for all training, validating and testing data. 

We trained the model based on a timeseries that represents a given interval of time, per aircraft, over a range of ${9}$ days.  In our experiment the default interval of time used is one hour. For each aircraft, we will randomly sample events every 5 minutes, per hour. The input feature vector for the autoencoder represents a fixed-length timeseries, for a given aircraft. This input feature vector will contain ${n}$ values, where ${n = x * r}$. Note, ${x}$ is the number of features in a single event and ${r}$ is the number of random samples selected for a given interval of time.

\subsubsection{Feature Scaling}
For most machine learning tasks the data pre-processing steps includes some form of feature value normalization. Techniques for normalization include encoding text values to numeric representations and feature scaling. Feature scaling ensures that large variations in feature values do not skew corresponding weights of features \cite{Sun2016}. 

Although the feature set for our experiment does not include text values, there are several numeric features that require normalization. There are two common approaches for feature scaling; one uses the min and max values of all values for each feature to calculate a normalized value and the other standardizes each feature value based on its mean and standard deviation.  Because there is the assumption that our data sets are normally distributed, we used the latter approach to standardized feature values using their corresponding mean and standard deviation. It should be noted that some features values such as longitude and latitude, require special consideration when calculating their mean and standard deviation.  For our purposes, we calculated simple arithmetic mean to scale location points. However, if we needed higher precision and our points were likely near the poles of the earth, the location point values should be converted to Cartesian coordinates before calculating their average and standard deviation.

\subsubsection{Average Vectors}
Once the average for each feature has have been calculated we will use these values to create a template, also know as an average vector for each type (i.e. aircraft). This vector will be be used in future work to calculate the variance between all patterns of the same type; which can be used to evaluate how effective our model will be at matching patterns for classification learning tasks. 

\section{Network Architecture}
For this experiment, we used a simple, LSTM based autoencoder architecture as tested by Suresh et al \cite{Suresh2017}. The autoencoder takes in an input feature vector that represents a fixed-length timeseries of 204 values (i.e. ${n = 17 * 12 = 204}$), for a given interval of time (i.e one hour). This feature vector is created by concatenating ${r}$ single event feature vectors together, for each hour. For flights whose duration is shorter than an hour, feature values will be zero padded. The single hidden (encoder) layer will take in ${m}$ values where ${m = 17}$, where 17 represents the compression factor of the autoencoder. The output layer outputs a vector of 204 values. The single hidden layer uses a ReLU activation function and the output layer uses a sigmoidal activation function.

\section{Results}
The primary objective of this experiment was to exploit the feature extraction capabilities of autoencoders to learn interesting features that could be later used to build a classifier for detecting surveillance aircraft using ADS-B flight data. Thus, we only used positive examples when training this model, so the standard metrics of loss, F1 etc are not applicable for this experiment. This choice allowed us to quickly learn features for these examples. The data set contains ${9017}$ examples of ${1}$ hour sequences of flight data, for ${41}$ unique aircraft. ${80\%}$ of the data set is dedicated to training and the other ${20\%}$ is dedicated to validation.

Of the 17 features listed in training examples, 10 of them were extracted as interesting features for the positive, surveillance aircraft examples. A number of these features such as longitude, latitude, speed and track were found by previous researchers to be important in their classification model, which was manually built based on their visual observations of flight patterns of surveillance aircraft \cite{defcon25spies}. \ref{table:2} contains the names of learned, extracted features.

\begin{table}[h!]
\begin{tabular}{ |m{1.5cm}|m{2cm}|m{4cm}| } 
 \hline
 Field & Data Type & Description \\
 \hline
  Alt & integer & The altitude in feet at standard pressure.\\
  \hline
  Lat & float & The aircraft’s latitude over the ground.\\
  \hline
  Long & float & The aircraft’s longitude over the ground.\\
  \hline
  PosTime & epoch (ms) & The time that the position was last reported by the aircraft.\\
  \hline
  Spd & knots (float) & The ground speed in knots.\\
  \hline
  Trak & degrees (float) & Aircraft’s track angle across the ground clockwise from 0° north.\\
  \hline
  Gnd & boolean & true, if aircraft is on the ground\\
  \hline
  Trt & integer & Transponder type.\\
  \hline
  Talt & number & The target altitude, in feet, set on the autopilot / FMS etc.\\
  \hline
  Ttrk & number & The track or heading currently set on the aircraft’s autopilot or FMS. \\
  \hline
\end{tabular}
\\
\caption{Learned Extracted Features \cite{adsbexchange}}
\label{table:2}
\end{table}

\section{Challenges}
The main challenges facing this experiment was due to the nature of the data itself. Not all surveillance aircraft fly on a regular basis, they have irregular flight patterns and irregular flight durations. All of these factors can vary greatly depending on the aircraft, law enforcement agency, jurisdiction etc. This can lead to poor data quality because we zero pad all missing data. In other words if the bulk of the needed duration's data is zero padded the examples are unusable after feature scaling and normalization.

These challenges made curating a usable data set into a tedious and time consuming process. In the future a more robust automated system must be implemented to gather training and test examples.

\section{Future Work}
The next steps are to test the features learned in this research by using them to train a classifier for identifying surveillance aircraft based solely on flight data. It should be noted that both negative and positive examples should be added to the new training data set. Future work also includes expanding experiments to test other methods of feature extraction to determine the best method for finding interesting features to be used to train more general aircraft classification models.

\bibliographystyle{IEEEtran}
\bibliography{main}

\begin{thebibliography}{1}
\providecommand{\url}[1]{#1}
\csname url@samestyle\endcsname
\providecommand{\newblock}{\relax}
\providecommand{\bibinfo}[2]{#2}
\providecommand{\BIBentrySTDinterwordspacing}{\spaceskip=0pt\relax}
\providecommand{\BIBentryALTinterwordstretchfactor}{4}
\providecommand{\BIBentryALTinterwordspacing}{\spaceskip=\fontdimen2\font plus
\BIBentryALTinterwordstretchfactor\fontdimen3\font minus
  \fontdimen4\font\relax}
\providecommand{\BIBforeignlanguage}[2]{{%
\expandafter\ifx\csname l@#1\endcsname\relax
\typeout{** WARNING: IEEEtran.bst: No hyphenation pattern has been}%
\typeout{** loaded for the language `#1'. Using the pattern for}%
\typeout{** the default language instead.}%
\else
\language=\csname l@#1\endcsname
\fi
#2}}
\providecommand{\BIBdecl}{\relax}
\BIBdecl

\bibitem{defcon25spies}
\BIBentryALTinterwordspacing
S.~Richards, J.~MacDonald-Evoy, and J.~Hernandez, ``{Spies in the Skies},''
  2017. [Online]. Available:
  \url{https://www.nstarpost.com/news/defcon-25-spies-in-the-skies/}
\BIBentrySTDinterwordspacing

\bibitem{adsbexchange}
\BIBentryALTinterwordspacing
``{Adsb Exchange}.'' [Online]. Available: \url{https://www.adsbexchange.com/}
\BIBentrySTDinterwordspacing

\bibitem{radar24}
\BIBentryALTinterwordspacing
``{Flightradar24}.'' [Online]. Available:
  \url{https://www.flightradar24.com/33.14,-90.54/7}
\BIBentrySTDinterwordspacing

\bibitem{Suresh2017}
\BIBentryALTinterwordspacing
H.~Suresh, P.~Szolovits, and M.~Ghassemi, ``{The Use of Autoencoders for
  Discovering Patient Phenotypes},'' no. Nips, 2017. [Online]. Available:
  \url{http://arxiv.org/abs/1703.07004}
\BIBentrySTDinterwordspacing

\bibitem{Sun2016}
J.~Sun, J.~Ellerbroek, and J.~Hoekstra, ``{Large-Scale Flight Phase
  Identification from ADS-B Data Using Machine Learning Methods},'' \emph{7th
  Int. Conf. Res. Air Transp.}, no. June, 2016.

\end{thebibliography}

\end{document}